\documentclass[letterpaper, 10 pt, conference]{ieeeconf} 
\IEEEoverridecommandlockouts      
\usepackage{cite}
\usepackage{amsmath,amssymb,amsfonts}
\usepackage{algorithmic}
\usepackage{graphicx}
\usepackage{textcomp}
\usepackage{xcolor}
\usepackage{caption}
\usepackage{subfig}
\usepackage{multirow}
\usepackage{hhline}
\usepackage{gensymb}
\usepackage{calrsfs}
\usepackage{leftidx}
\usepackage{mathtools}
\usepackage{comment}
\usepackage{amsmath}
\usepackage{url}
\usepackage{hyperref}
\usepackage{float}
 
\DeclareMathOperator*{\argmin}{argmin}
\def\BibTeX{{\rm B\kern-.05em{\sc i\kern-.025em b}\kern-.08em
    T\kern-.1667em\lower.7ex\hbox{E}\kern-.125emX}}

\title{\LARGE \bf
Localization and Offline Mapping of High-Voltage Substations in Rough Terrain Using a Ground Vehicle
}

\author{}
\author{Ioannis Alamanos$^{*}$, George P. Moustris and Costas S. Tzafestas
\thanks{All authors are with the School of Electrical \& Computer Engineering, National Technical
University of Athens, Greece, Corresp. author email:
$^{*}${\tt\small ioannesalamanos@gmail.com}}%
}

\begin{document}

\maketitle

\begin{abstract}
This paper proposes an efficient hybrid localization framework for the autonomous navigation of an unmanned ground vehicle in uneven or rough terrain, as well as techniques for detailed processing of 3D point cloud data. The framework is an extended version of FAST-LIO2 algorithm aiming at robust localization in known point cloud maps using Lidar and inertial data. The system is based on a hybrid scheme which allows the robot to not only localize in a pre-built map, but concurrently perform simultaneous localization and mapping to explore unknown scenes, and build extended maps aligned with the existing map. Our framework has been developed for the task of autonomous ground inspection of high-voltage electrical substations residing in rough terrain. We present the application of our algorithm in field trials, using a pre-built map of the substation, but also analyze techniques that aim to isolate the ground and its traversable regions, to allow the robot to approach points of interest within the map and perform inspection tasks using visual and thermal data.
\end{abstract}

\section{Introduction}
The localization problem refers to the pose estimation of a mobile robot within a prior map. This research topic, although necessary for autonomous navigation of robots in a known environment, it has not attracted enough attention throughout the last years, especially in the case of outdoor navigation in uneven or rough terrain, which requires the use of the three-dimensional information of the surroundings, and not just a 2D slice. For this reason, most of the currently used localization algorithms are just wrappers of well-known simultaneous localization and mapping (SLAM) algorithms which emphasize the incremental creation of a 3D map with concurrent estimation of the pose of the robot. Besides, these extended versions of the SLAM algorithms focus on the exclusive localization within a known map. However, in a dynamic environment where the map continually changes, such a system would fail to localize as it would not find enough correspondences between 3D points, or present many mismatches. Only if the algorithm updates the map, can it accurately localize within it, with reference to the prior map. 

\begin{figure}[tbh] 
  \centering
  \includegraphics[width=\columnwidth]{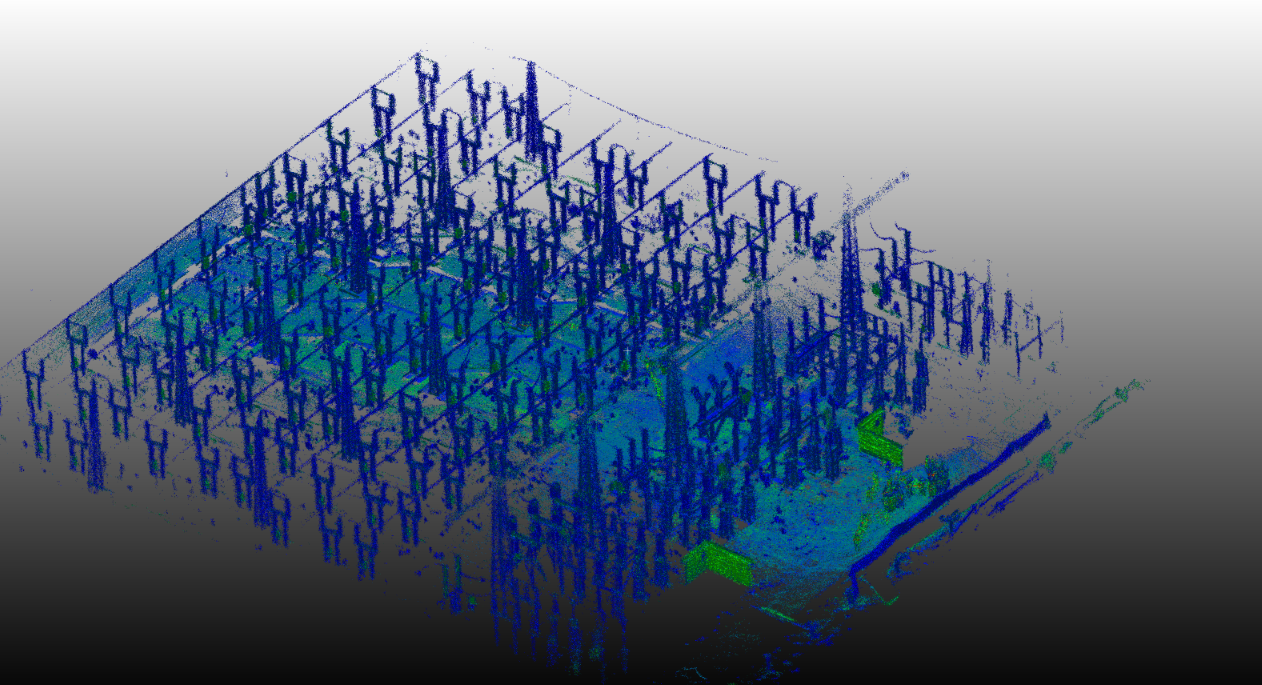}
  \caption{Generated Map of HVSS.}
\label{fig:Total_Map}
\end{figure}

Crafting a utilizable 3D point cloud is a crucial task. Every current Lidar-inertial odometry algorithm builds a noisy 3D point cloud. The noise comes from errors of the Lidar measurements, the motion undistortion from the Inertial Measurement Unit (IMU) measurements, as well as the misregistration of points. As a result the surfaces of the objects can have an undesirable thickness which hinders crucial processes, such as raycasting, used to determine the visibility region of points of interest, or even accurately define then within the map. Evidently, generating a sharp de-noised 3D model is necessary to perform inspection tasks within a known environment.

After acquiring an accurate point cloud and determining the regions of interest, the next task is to successfully navigate within the map of the high-voltage electric substation (HVSS). To form a safe and feasible path, the traversable regions of the ground must be established. The precise extraction of the ground in a point cloud is not a widely researched topic. Almost every algorithm is based on the cloth simulation filter (CSF) (e.g. \cite{2,3}) which takes the inverted surface and places above cloth nodes which interact with the points that belong to the ground. Another approach is the three dimensional random sample consensus (RANSAC) \cite{4}; however it would only be successful if the ground plane is the dominant plane of the point cloud.

The topic of modeling a point cloud of the ground to obtain information regarding traversability has been widely investigated in the last years. Implementations which take advantage of the normals of the point cloud have been developed to estimate safe regions, even in real-time, with significant accuracy.

This work describes a complete pipeline for crafting high-quality maps from collected point clouds and navigating within a rough terrain inside an HVSS using ground vehicles. The main contributions of this paper are:
\begin{itemize}
  \item An extended version of FAST-LIO2 able to localize within a known environment and even update the pre-built map through a hybrid scheme.
  \item Techniques to smooth and de-noise 3D point clouds generated from real-time SLAM algorithms and extract the ground map.
  \item Application of methods to determine the traversable regions of rough ground terrain for safe navigation within a HVSS.
\end{itemize}

For the benefit of the community we have made the code open source and can be found on \url{https://github.com/iral-ntua/FAST_LIO_LOCALIZATION}.

\section{Related Work}
\label{chap:RW}
In recent years, the development of algorithms focusing exclusively on localization within a known 3D point cloud has been limited, with one notable exception being the 3D Adaptive Monte Carlo Localization (AMCL) \cite{5}. This algorithm serves as an extension of the widely employed 2D AMCL, employing a probabilistic approach with a particle filter. However, its efficiency is contingent on the availability of precise odometry data to serve as an initial guess for the optimization problem. Despite its computational efficiency, obtaining accurate odometric information in 3D often necessitates the use of SLAM algorithms, which can be computationally demanding.

Conventional localization methods often rely on iterative closest point (ICP) \cite{6} for aligning each Lidar scan with the prior map. However, these methods can struggle, particularly with large point clouds, as they may not meet real-time requirements. In contrast, recent 3D SLAM algorithms utilizing Lidar sensors have proven both computationally efficient and accurate, alleviating much of the computational burden. One notable example is the LIORF module based on the LIO-SAM framework \cite{7}, which takes a prior map as input. Another wrapper involves an extended version of FAST-LIO2, incorporating a module that asynchronously performs ICP on the known map, adjusting the pose and the map to the prior. However, the continuous execution of traditional ICP remains computationally demanding for real-time applications.

To address the challenge of generating a reliable point cloud, various techniques have been devised for point cloud de-noising. Many traditional approaches, rooted in image processing, formulate an optimization problem by leveraging the geometric attributes of the point cloud, such as normals, while discarding outliers. Among these, the \textit{bilateral filter} \cite{8} stands out. This method posits that the de-noised point cloud results from the original point cloud with a displacement factor for each point. This factor is derived from point normals via principal component analysis and Gaussian weights. However, adjusting the parameters of these weights can sometimes be challenging for users aiming to achieve desirable results.

Another notable method is the \textit{guided filter} \cite{9}, which assumes a linear local model around each 3D point. It tackles the de-noising task by solving a least squares problem to minimize the reconstruction residual. The guided filter excels in preserving edges and sharp shapes, distinguishing itself from other smoothing filters. Nevertheless, it exhibits less tolerance towards noisy point clouds, as it is inherently tied to explicit normal estimation.

Recent advancements in point cloud de-noising involve the integration of deep learning-based models. For instance, the Neural Projection Denoising algorithm (NPD) \cite{10} estimates reference planes for sets of points by computing the normal vector for each point using both local and global information. This approach enhances robustness to noise intensity and curvature variation. Another innovative approach is the Total Denoising Neural Network \cite{11}, which focuses on unsupervised learning, utilizing only noisy point cloud data for training. Despite its unsupervised nature, this method produces comparable experimental results to supervised techniques.

Proper modeling of the ground is a critical task for both wheeled and legged robots, as it forms an integral part of the core components, along with the localization module, for ensuring the safe navigation of Unmanned Ground Vehicles (UGV). A notable work that stands out for its real-time capabilities is presented in \cite{16}. This approach employs a Bayesian generalized kernel inference method based on the work of Vega-Brown et al. \cite{17}. The method involves assigning grid cells to the point cloud and executing the Bayesian generalized kernel inference in two steps. First, a regression is performed to obtain a dense elevation map, and second, a classification is carried out to determine traversable regions.

An extension of this algorithm is discussed in \cite{18}, which introduces additional functionality for semantic segmentation using visual information from a camera. Segmenting the map based on visual cues to identify traversable regions becomes crucial, especially as the algorithm, by estimating the roughness of the terrain, may sometimes overlook movable objects (such as small plants) or impassable obstacles.

\section{System Overview}
\begin{figure}[tbh]
  \centering
  \includegraphics[width=0.9\columnwidth,]{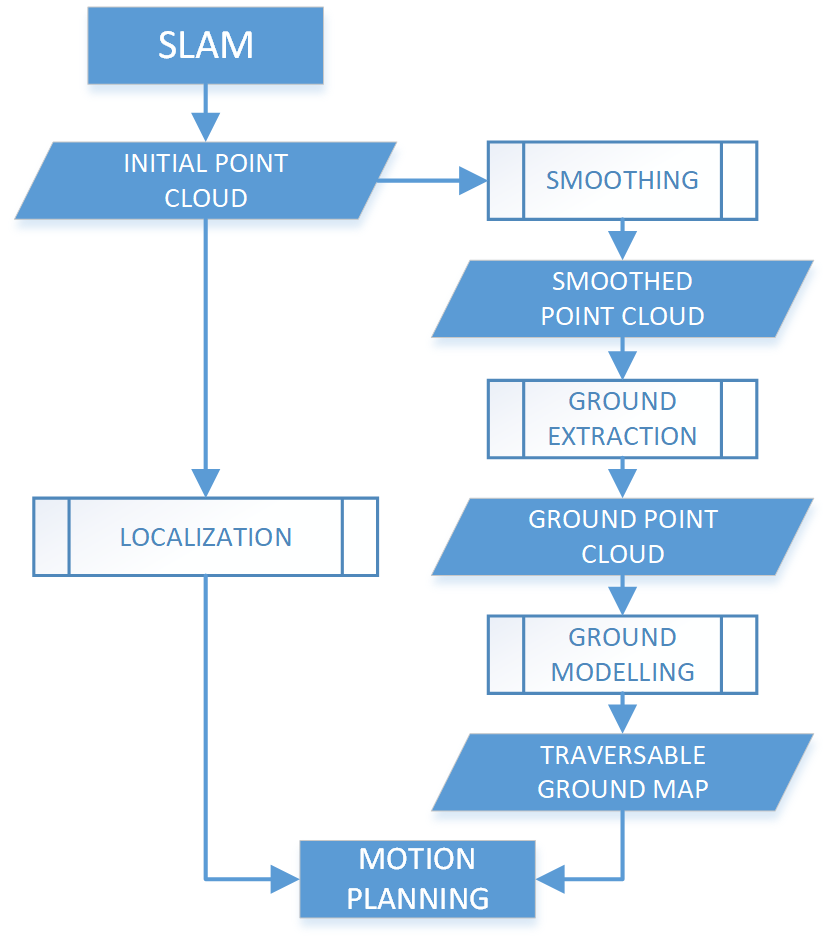}
  \caption{Overview of the general pipeline of the system.}
  \label{fig:f1}
\end{figure}

The proposed system overview is depicted in Fig. \ref{fig:f1}. The initial map of the HVSS is generated by the SLAM algorithm, resulting in a point cloud which further undergoes a smoothing post-processing step. Subsequently, a ground extraction technique is applied to separate the terrain, forming the basis for constructing the traversability map. This map becomes crucial for planning the motion of the robot, enabling it to explore specific user-defined regions of interest. Concurrently, taking as input the SLAM-generated point cloud, a localization module estimates the pose of the UGV within the prior map. This localization information is essential for executing the motion commands generated by the planner. The proposed scheme functions as a subsystem within the broader system, specifically designed for inspection purposes.

\section{Method}

\subsection{Localization}

Our localization framework is based on the architecture of FAST-LIO2 \cite{1}. Unlike most of Lidar SLAM algorithms, FAST-LIO2 is computationally more efficient and precise due to its novel functionalities. Its basic pipeline for pose estimation and 3D point cloud creation is the following:

\begin{enumerate}
\item Scan-to-scan motion undistortion of the raw Lidar points through back-propagation by using the measurements from an inertial measurement unit (IMU). 
\item Pose estimation on each scan. To estimate the pose of the UGV on each scan, a non-linear least squares problem is formulated to minimize the residuals resulting from the registration of the current scan points with those of the map. While many methods commonly employ techniques such as Levenberg-Marquardt or Gauss-Newton, FAST-LIO2 employs an Extended Kalman Filter (EKF). In the state prediction and covariance calculation steps, the algorithm utilizes IMU measurements between Lidar scans. During the update step, the residuals from point-to-plane registrations are minimized. The algorithm incorporates a novel incremental kd-tree (ikd tree) \cite{19} that demonstrates good performance for k-nearest neighbors searches in Lidar odometry. In practice, the algorithm forms a plane from the five nearest neighbors of each current Lidar scan point on the map. If the point-to-plane distance falls below a predefined threshold, it registers this point to the plane formed by its neighbors. This plane is essentially described by its centroid and point normals. By adopting this approach, FAST-LIO2 effectively refines the UGV's pose estimation, demonstrating a balance between computational efficiency and accurate registration.
\item After the state estimation, the method updates the overall point cloud and its ikd-tree with the new odometry-registered scan.
\end{enumerate}

The contribution of our extended version of FAST-LIO2 can be summarized as follows:

\begin{enumerate}
\item Robust localization on pre-built maps.
\item Pose initialization within the known map using an initial guess and ICP.
\item Publication of complete odometry messages for real-world applications.
\item Improved memory handling regarding Lidar and IMU data in case deprecated messages are received.

\end{enumerate}

\begin{figure}[tb] 
  \centering
    \includegraphics[width=\columnwidth ]{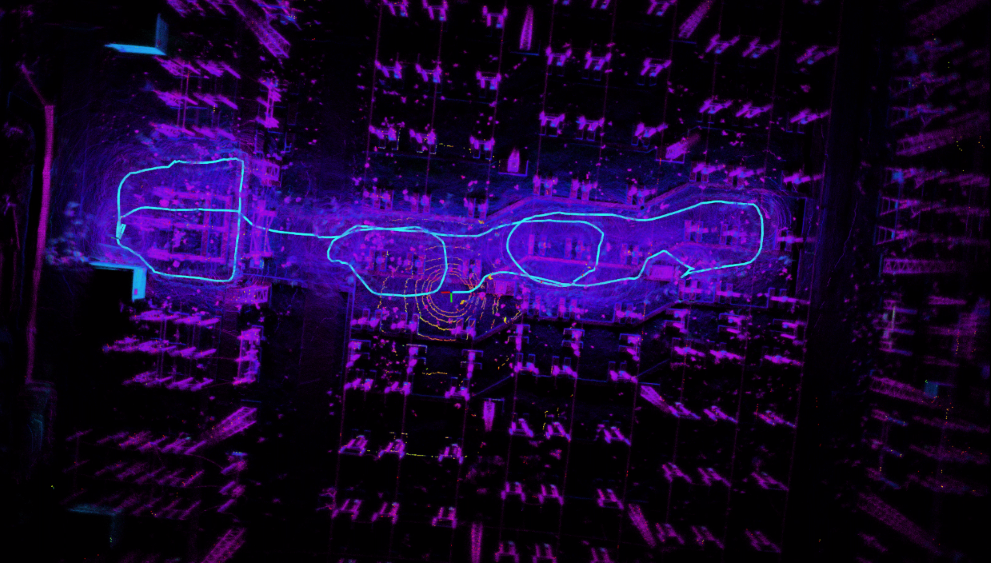}
  \caption{Application of our method within a HVSS.}
\label{fig:LOC}
\end{figure}

Our localization method maintains the core of FAST-LIO2 regarding motion undistoriton and state estimation. Nevertheless, the ikd-tree of the overall active map that is used for each scan's registration, is not formed and updated incrementally from each scan but is loaded from a prior point cloud that serves as the initial map. The reason for using the ikd-tree and not a static kd-tree is that the user has the ability to update the tree and essentially perform SLAM with a prior map by assigning the desired time to a parameter. This allows the robot to localize in a known environment and sequentially explore a new scene, generating an updated point cloud aligned with the prior one. To the best of our knowledge, there is no other hybrid SLAM-Localization algorithm that can perform in such scenarios. The only required input from the user is the prior point cloud and an initial guess of the pose of the robot relative to the frame of the map. Relevant operation of our method is demonstrated in Fig. \ref{fig:LOC}.

One of the main challenges of solving the localization problem is to estimate the initial pose of the robot within the prior map. Given two sets of points $P = \{p_{1},..., p_{N}\}, Q = \{q_{1},...,q_{M}\} \in \mathbb{R}^3$, we seek to optimize a rigid transformation matrix $T \in \mathbb{R} ^{4 \times 4}$, comprised of a rotation matrix $R \in \mathbb{R} ^{3 \times 3}$ and a translation vector $t \in \mathbb{R} ^3$, in order to align P with Q.
\begin{equation}
\pmb{T}^* = \argmin_{\pmb{R},\pmb{t}}\sum_{i = 1}^{N}||\pmb{R}p_{i}+\pmb{t}-\hat{q_{i}}||^2 + I_{SO(d)}(\pmb{R}),
\label{eq:1}
\end{equation}
where $\hat{q_{i}} \in Q$ is the corresponding point of $p_{i}$, $||\pmb{R}p_{i}+\pmb{t}-\hat{q_{i}}||$
is the distance from the transformed source point to the target  corresponding point, and $I_{SO(d)}(\pmb{R})$  is an indicator function for the special orthogonal group $SO(d)$, which requires $\pmb{R}$ to be a rotation matrix:
\begin{equation}
I_{SO(d)}(\pmb{R}) = \begin{cases}
0, \text{ if } \pmb{R}^T \pmb{R} = \mathbf{I} \text{ and } det({\pmb{R}}) = 1, \\
+ \infty, \text{ otherwise}.
\end{cases}
\label{eq:2}
\end{equation}
To estimate the desired transformation matrix $\pmb{T}^*$, first we accumulate the first ten scans (source point cloud) while keeping the Lidar static, and then perform ICP between these scans and the prior map (target point cloud) using the PCL library \cite{20}. The ICP method uses an iterative approach, and assuming that we are on the $k_{th}$ iteration, the problem is solved in the two following steps:
\begin{enumerate}
\item \textit{Corresponding points}: find the closest point $\hat{q_{i}}^{(k)} \in Q$ for each transformed point $p_{i}$:
\begin{equation}
\hat{q_{i}}^{(k)} = \argmin_{q \in Q}||R^{(k)}p_{i}+t^{(k)}-q||.
\label{eq:3}
\end{equation}
\item \textit{Transformation update}: optimize the transformation matrix by minimizing the 3D euclidean distance between the corresponding sets of points:
\begin{equation}
\begin{multlined}
\pmb{T}^{(k+1)} = \\
\argmin_{\pmb{R}^{(k)},\pmb{t}^{(k)}}\sum_{i = 1}^{N}||\pmb{R}^{(k)}p_{i}+\pmb{t}^{(k)}-\hat{q_{i}}^{(k)}||^2 + I_{SO(d)}(\pmb{R}).
\end{multlined}
\label{eq:4}
\end{equation}
\end{enumerate}

To iteratively determine the optimized transformation matrix, ICP solves equation (\ref{eq:4}) in closed form via Singular Value Decomposition \cite{svd}. The drawback of this method is that an initial guess of the robot pose is needed for the ICP to converge and accurately localize the robot to the correct map position. Considering this constraint, except for the convergence of the ICP, to ensure an exact initialization within the map, we take advantage of a Euclidean fitness score which expresses the mean of squared distances from the source to the target as shown in equation (\ref{eq:5}). Here $x_{i}$ and $\hat{x_{i}}$ are corresponding points of the source and target clouds. The pose initialization is considered successful only if the Euclidean fitness score is below a predefined threshold, set to 0.01 in our case.

\begin{equation}
FS =  \frac{\sum_{i = 1}^{N}  (x_{i} - \hat{x_{i}})^2}{N}
\label{eq:5}
\end{equation}

\subsection{Map Crafting}

\begin{figure}[tbh] 
  \centering
    \includegraphics[width=\columnwidth]{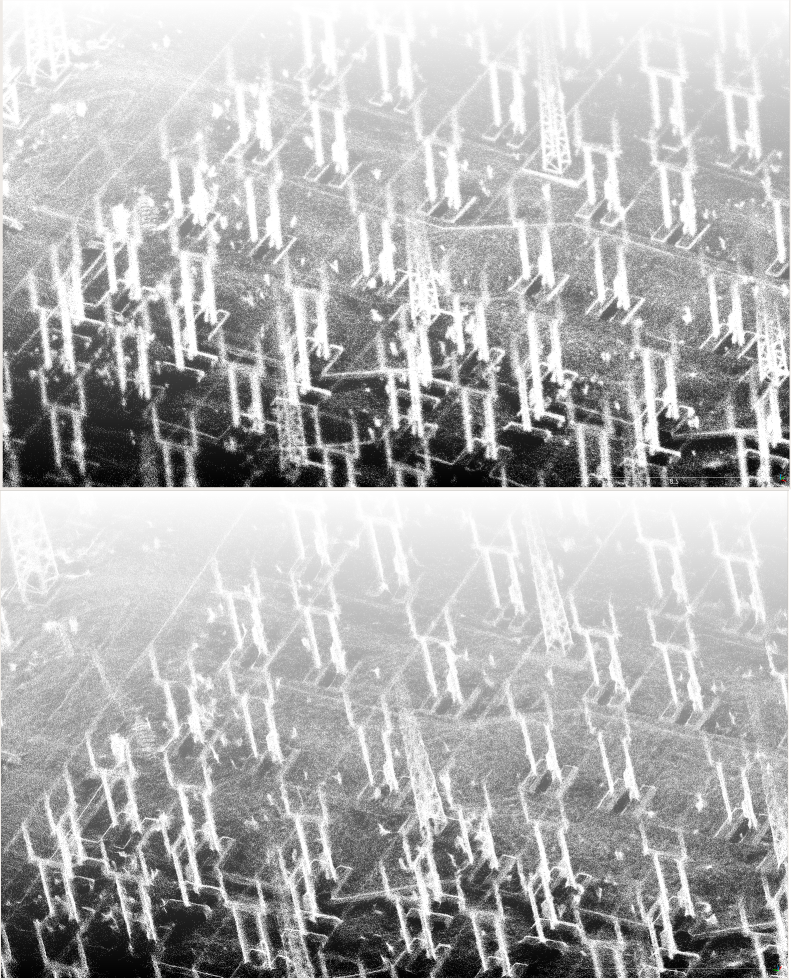}
  \caption{Raw and filtered part of map (upper and lower correspondingly).}
\label{fig:Filt}
\end{figure}

To obtain an initial 3D point cloud we used the FAST-LIO2 SLAM algorithm. However the resulting map, which is formulated by accumulated scans registered to the odometry frame, is not usable as it contains considerable amount of noise due to sensor and algorithmic imperfections. Thus, there is an undesirable thickness to every surface of the point cloud which renders it impossible to utilize some core methods needed for the autonomous inspection of the HVSS, e.g.raycasting between points of interest and the ground. To address this problem and de-noise the point cloud we perform a two step filtering procedure:

\begin{enumerate}
    \item First, we apply a uniform sampling filter to assign voxels to the 3D continuous space. For each point we only keep the voxel that is closer to it, if that exists. By uniformly discretizing the three dimensional space, not only we remove noise but also keep only necessary meaningful information by making the point cloud more sparse and with specified structure.
    \item Following, we use the Moving Least Squares (MLS) \cite{21} algorithm of PCL library to de-noise and smooth the point cloud. To define surfaces (planes) through sets of points, MLS minimizes the weighted least square error which best fits the points to a surface. Specifically, the problem can be solved by minimizing the following term:
\begin{equation}
\sum_{i = 1}^{N}  (\langle n_{i}p_{i}\rangle-D)^2 \theta(||p_{i}-q||),
\label{eq:6}
\end{equation}

where the local plane is defined as: $\kappa = {x|\langle n_{i}p_{i}\rangle -D=0, x \in \mathbb{R} ^3} , n \in \mathbb{R} ^3, ||n||=1$, $p_{i}$ is a 3D point, $n_{i}$ its corresponding normal, $D$ the plane model, q the projection onto the plane, and $\theta$ a smooth monotonous decreasing function. By aligning every point set to its local surface, the noise from the point cloud diminishes and concurrently the normal computation of the points is enhanced. The main drawback of this method is that sharp edges within the point cloud are slightly smoothed. Nevertheless, as demonstrated in Fig. \ref{fig:Filt}, this two-step filtering yields significantly enhanced and denoised results.
\end{enumerate} 

\subsection{Traversability Mapping}

\begin{figure}[tbh] 
  \centering
    \includegraphics[width=\columnwidth, height=8.2cm]{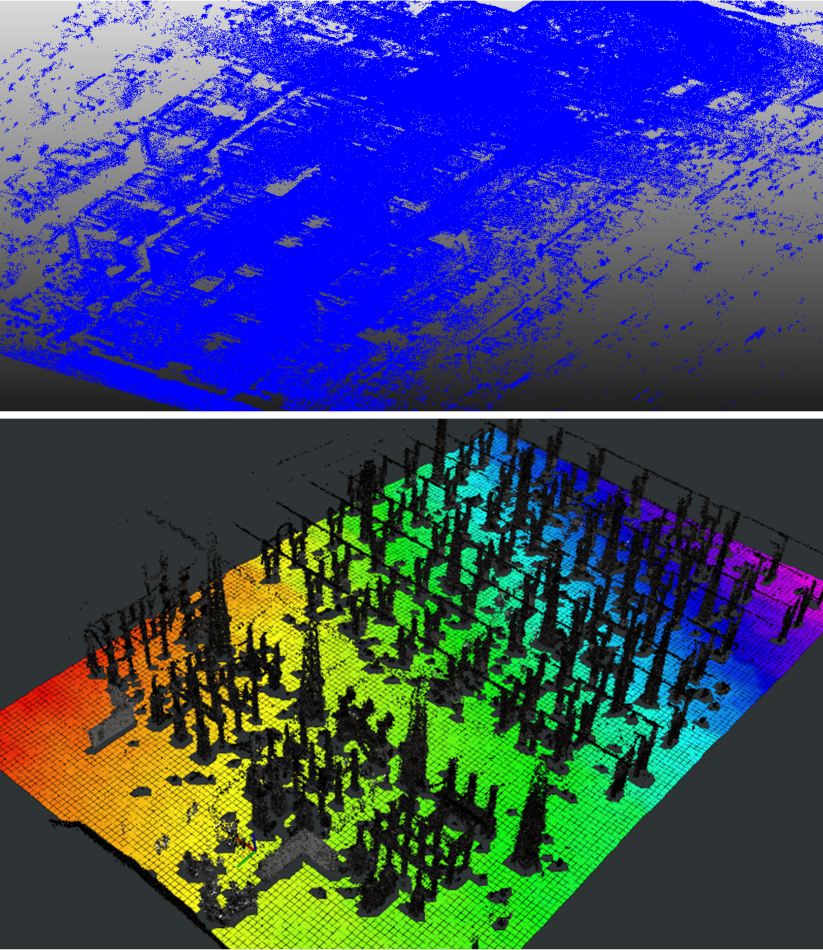}
  \caption{(UP) Ground point cloud extracted from CSF. (DOWN) Grid map with overground structures.}
\label{fig:ground}
\end{figure}

Defining the traversable regions of the rough terrain of a HVSS is essential for the safe navigation of the UGV. For the precise modeling of the ground, the input point cloud needs to be noise-free and the ground meticulously divided from the overground components. To attain this, we first utilize the previously described method to acquire a 'clean' point cloud with carefully computed point normals; following, we use a CSF filter with appropriate parametrization of cloth resolution and classification threshold, which expresses the size of each cluster of the map that will be classified as \textit{ground} or \textit{non-ground}, to isolate the terrain (Fig. \ref{fig:ground}).

To this end, we utilized the Grid Map open-source library \cite{22}, which focuses on online surface reconstruction and interpretation of rough terrain. Through this library, we converted the terrain point cloud to a Grid Map object, essentially interpreting the ground as a grid of cells, assigning to each one a value expressing its \textit{elevation} (Fig. \ref{fig:ground}). To determine the traversable ground regions, three more filters were applied to corresponding layers,
\begin{enumerate}
    \item Surface Normals Filter: Estimates the normal of each cell and is vital for the operation of the next filters.
    \item Slope Filter: Calculates the slope of each cell by directly taking advantage of the surface normals.
    \item Roughness Filter: Computes the roughness of each cell by utilizing the information about the  normals surrounding that cell. 
\end{enumerate} 
By normalizing the values of the \textit{Slope} and \textit{Roughness} layers and using appropriate thresholds to reflect the capability of the ground vehicle to pass through rough terrain, we can assign a cost to each cell according to its \textit{traversability} and thus specify traversable regions. This produces a 2D costmap which can be used in ROS for subsequent navigation.

\section{Experimental Results}

\subsection{Metric and Experimental Setup}
In this section we assess the performance of our localization method in comparison to LIORF and traditional ICP localization. Given that there is no ground-truth for the evaluation, the metric used for the comparison of the algorithms is the \textit{mean cloud-to-cloud distance} of the point clouds generated using the localization algorithms and the prior map. Taking into account that the localization point clouds are formed from odometry-registered undistorted scans and that the purpose of a localization algorithm is the accurate positioning of the robot within a known map, cloud-to-cloud distance is a representative metric for the performance of the methods. Cloud-to-cloud distance is computed by determining corresponding closest points between the two point clouds as expressed in equation (\ref{eq:3}), and computing the mean error between these point sets (same as in equation (\ref{eq:5})). Rejecting as outliers corresponding points whose distance is above from a predefined threshold, is a key step as new scenes maybe have been mapped during localization and would adulterate the final results.

All the experiments have been conducted on a laptop computer with an Intel Core i9-9980 CPU and 32 GB RAM, and the data have been collecting using Robosense's RS-LiDAR-16 and Vectornav's VN-100 IMU. We did not perform any further processing time evaluation, since the core of the method remains the same as in FAST-LIO2.

\subsection{Evaluation}
For the evaluation of the algorithms we collected two distinct datasets from the HVSS. From the first dataset we generated the prior map through FAST-LIO2, while the second was used to assess the localization methods. The experimental results are presented in Table \ref{tab:1}.

\begin{figure}[tbh] 
  \centering
    \includegraphics[width=\columnwidth, , height=14.3cm]{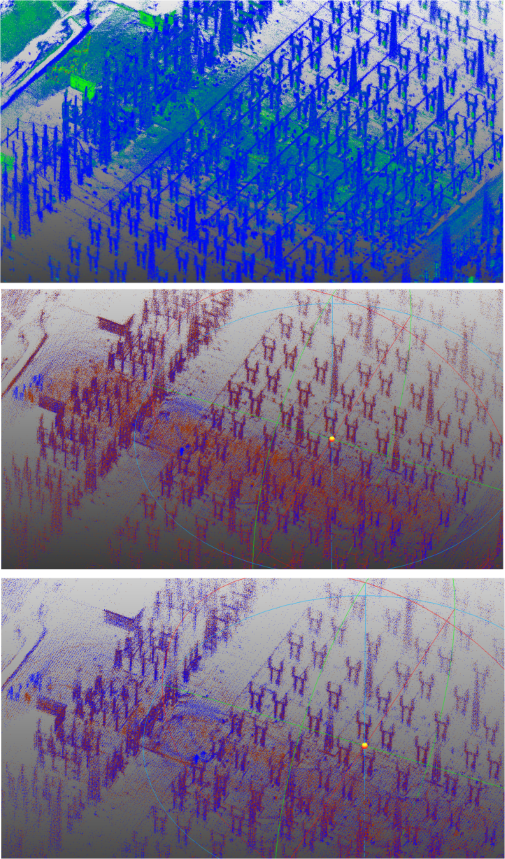} 
  \caption{(UP) Raw prior map; (MIDDLE) prior map (Blue) along with point cloud from LIORF (Red); (BOTTOM) prior map (Blue) along with point cloud from FAST-LIO-LOCALIZATION (Red).}
\label{fig:C-C}
\end{figure}

\begin{center}
\begin{table}[h]
\caption{Performance comparison of localization methods}
\begin{center}
\begin{tabular}{ |l|c|c| } 
\hline
Method & Mean Error (m) & Standard Deviation (m) \\ 
\hline
ICP &  - & - \\ 
\hline
LIORF & 0.050 & 0.069 \\ 
\hline
FAST-LIO-LOC. & \textbf{0.026} & \textbf{0.049} \\ 
\hline
\end{tabular}
\label{tab:1}
\end{center}
\end{table}
\end{center}
From the obtained results, it is evident that our method outperforms the others in terms of accuracy. Traditional ICP fails to localize the sensor after a few seconds due to its high computational burden. Although Lidar operates at 10hz the algorithm is able to perform at 2Hz at most, on an i9 cpu. On the other hand, LIORF is much more effective and localizes the robot during the whole dataset with low drift. However, it is considerably demanding on computational resources. In contrast, our method is more precise in terms of accuracy, but also significantly more lightweight as presented on \cite{1} (comparison of FAST-LIO2 with LIO-SAM). The alignment of the corresponding point clouds is presented in Fig. \ref{fig:C-C}. As is evident in the middle depiction, the red point cloud from LIORF is much more dominant due to the existing noise around the surfaces, even though significantly more sparse, as only the scans from the keyframes are projected to the map. On the other hand, on the bottom picture where every scan from FAST-LIO-LOCALIZATION is aligned to the map, the colors are balanced as the points lie almost exactly on the same positions relative to the prior map.

\section{Conclusion}

This paper proposed a hybrid localization-SLAM method, and an effective scheme for filtering raw point clouds to generate noise-free maps. Our localization method is based on the framework of FAST-LIO2, while map crafting incorporates a two step filtering including a uniform sampling filter, and a smoothing filter which utilizes moving least squares algorithm. The fabrication of a noise-free point cloud enables the efficient development of essential tasks for the inspection of points of interest within the HVSS, and the localization module is necessary for the safe navigation of the UGV. The demonstrated material corroborates the effectiveness of our map crafting method, while the presented results related to localization confirm the superiority of our method in comparison with the current most robust localization algorithms.

\bibliographystyle{unsrt} 
\bibliography{refs} 

\end{document}